%% file: root.tex
\documentclass[letterpaper, conference]{ieeeconf}
\IEEEoverridecommandlockouts   
\usepackage{amsmath,amsfonts}
\usepackage{algorithmic}
\usepackage{algorithm}
\usepackage{array}
\usepackage[caption=false,font=normalsize,labelfont=sf,textfont=sf]{subfig}
\usepackage{textcomp}
\usepackage{stfloats}
\usepackage{url}
\usepackage{verbatim}
\usepackage{graphicx}
\usepackage{cite}
\usepackage[colorlinks,linkcolor=blue]{hyperref} 
\usepackage{mathrsfs}
\usepackage[flushleft]{threeparttable}
\usepackage{tablefootnote}
\usepackage{multirow}
\usepackage{graphicx}
\usepackage{color}
\usepackage{tikz}
\usetikzlibrary{calc,arrows,decorations.markings}
\usepackage{balance}
\usepackage{booktabs}
\usepackage{xpatch}
\usepackage{mathrsfs}
\usepackage{float}

\makeatletter
\usepackage{etoolbox,lipsum}
\patchcmd\@makecaption{\\}{.~}{}{\fail}

\graphicspath{{./imgs/}}

\usepackage{caption}

\title{\LARGE \bf Grasp as You Dream: Imitating Functional Grasping from \\ Generated Human Demonstrations}

\author{
Chao Tang, Jiacheng Xu, Haofei Lu, Bolin Zou, Wenlong Dong, Hong Zhang, and Danica Kragic
\thanks{Chao Tang, Jiacheng Xu, Haofei Lu, and Danica Kragic are with the Department of Robotics, Perception and Learning, KTH Royal Institute of Technology, Stockholm, Sweden.}
\thanks{Bolin Zou, Wenlong Dong, and Hong Zhang are with the Shenzhen Key Laboratory of Robotics and Computer Vision, Southern University of Science and Technology, Shenzhen, China.}
}

\begin{document}

\maketitle
\thispagestyle{empty}
\pagestyle{empty}

\input{abstract}

\input{intro}

\input{related}

\input{approach}

\input{exp_setup}

\input{exp}

\input{discussion}

\bibliographystyle{IEEEtran}
\bibliography{root}

\newpage

\end{document}

%% file: abstract.tex
\begin{abstract}
Building generalist robots capable of performing functional grasping in everyday, open-world environments remains a significant challenge due to the vast diversity of objects and tasks. Existing methods are either constrained to narrow object/task sets or rely on prohibitively large-scale data collection to capture real-world variability. In this work, we present an alternative approach, GraspDreamer, a method that leverages human demonstrations synthesized by visual generative models (VGMs) (e.g., video generation models) to enable zero-shot functional grasping without labor-intensive data collection. The key idea is that VGMs pre-trained on internet-scale human data implicitly encode generalized priors about how humans interact with the physical world, which can be combined with embodiment-specific action optimization to enable functional grasping with minimal effort. Extensive experiments on the public benchmarks with different robot hands demonstrate the superior data efficiency and generalization performance of GraspDreamer compared to previous methods. Real-world evaluations further validate the effectiveness on real robots. Additionally, we showcase that GraspDreamer can (1) be naturally extended to downstream manipulation tasks, and (2) can generate data to support visuomotor policy learning. Project webpage can be found \href{https://sites.google.com/view/graspdreamer/home}{here}.

\end{abstract}

%% file: intro.tex
\section{Introduction}
Human grasping behavior encodes rich functional constraints, revealing where and how objects should be grasped to enable their intended use, beyond merely being lifted. This perspective naturally leads to the concept of functional grasping \cite{kokic2017affordance,song2013predicting}. However, it remains non-trivial for robots to perform functional grasping in general due to the diversity of objects and tasks in everyday, open-world environments

\begin{figure}[t]
  \centering
  \begin{tikzpicture}[inner sep = 0pt, outer sep = 0pt]
    \node[anchor=south west] (fnC) at (0in,0in)
      {\includegraphics[height=4.8in,clip=true,trim=0.1in 0in 0in 0in]{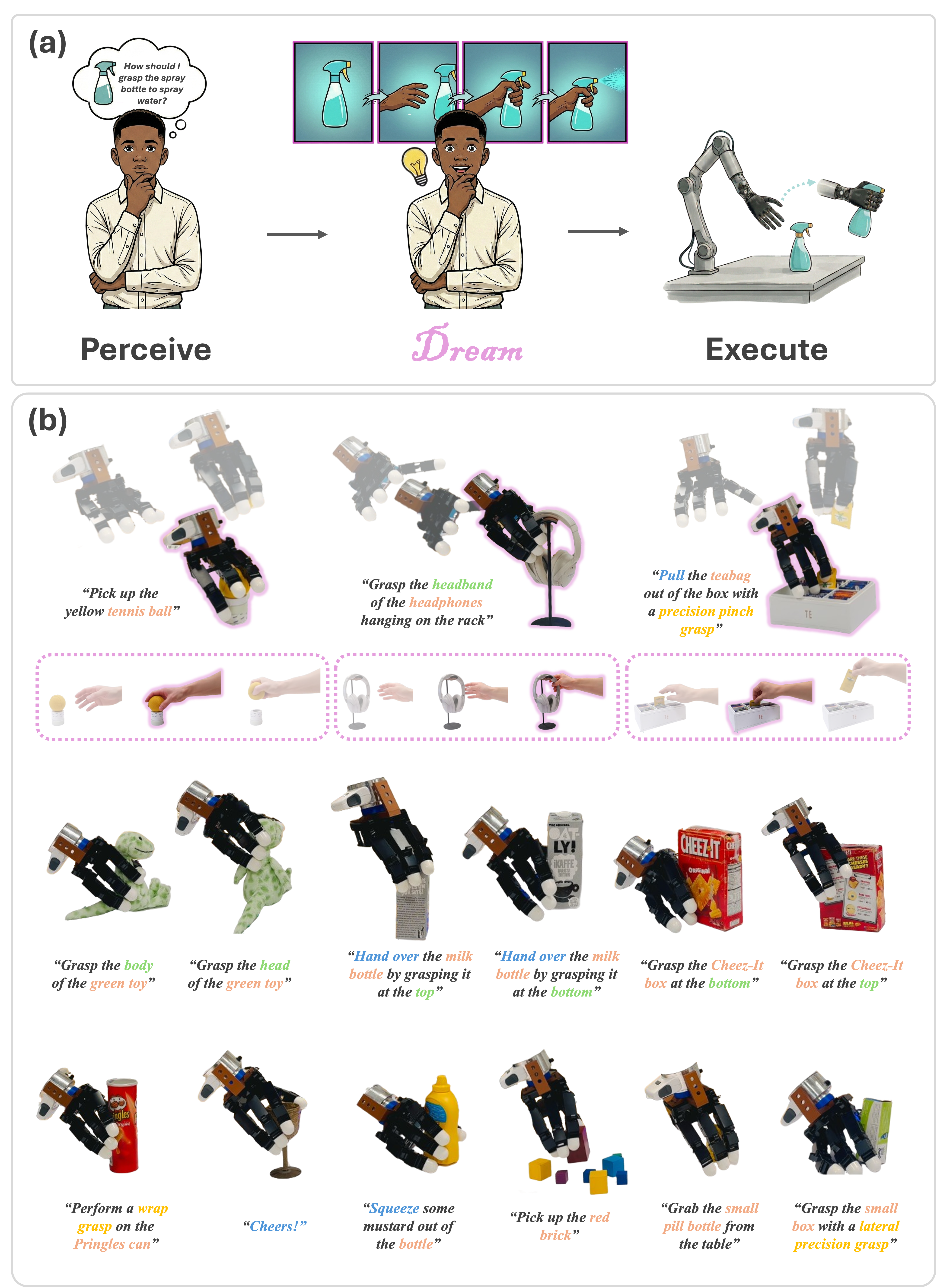}};
  \end{tikzpicture}
        \caption{GraspDreamer leverages human demonstrations synthesized by visual generative models (VGMs), such as video generation models, to enable zero-shot functional grasping. (a) Conceptual illustration of GraspDreamer. (b) Example generated human demonstrations and the corresponding executed dexterous functional grasps.}
  \label{fig:concept}
  \vspace*{-0.3in}
\end{figure}

Recently, data-driven approaches \cite{weng2024dexdiffuser,deshpande2025graspmolmo,huang2025hgdiffuser}, trained on curated functional grasping datasets \cite{murali2021same,wang2023dexgraspnet,yang2022oakink}, have achieved strong performance on closed-set benchmarks.
However, their ability to generalize remains limited in open-world settings, where object geometry, appearance, and task intent vary widely. To expand data diversity and coverage, recent works incorporate foundation models into their pipelines. These methods either fine-tune foundation models for functional grasping and manipulation \cite{he2025dexvlg,deng2025graspvla,deshpande2025graspmolmo}, or directly leverage pre-trained models to provide grasp-related knowledge \cite{tang2023graspgpt,tang2025foundationgrasp}. Nevertheless, both directions often rely on large-scale data collection to bridge the gap between generic foundation-model priors and executable robot actions.

In this work, we propose an alternative approach, \textbf{GraspDreamer}, as illustrated in Figure \ref{fig:concept}. Instead of collecting large-scale functional grasping datasets as in prior work, we leverage recent visual generative models (VGMs) (e.g., Veo) pre-trained on internet-scale data as a scalable source of open-world interaction priors for zero-shot functional grasping. Our key insight is that VGMs capture semantically grounded, task-oriented motion priors that can be coupled with embodiment-specific action optimization to generate physically executable functional grasps with minimal effort. Specifically, GraspDreamer consists of three stages. In the first stage, it extracts task-relevant cues from user inputs and accordingly generates human demonstrations that reflect the desired functional intent. In the following stage, GraspDreamer extracts human hand trajectories from the generated demonstrations and refines them via the hand trajectory optimization, yielding temporally coherent and physically plausible hand motions. Finally, GraspDreamer transfers human motions to the target robot hand through human-to-robot (H2R) functional retargeting, including VLM-based hand affordance reasoning, taxonomy-aware kinematic retargeting, and hand-object contact refinement, to generate executable functional grasp configurations for deployment. Extensive experiments on the public functional grasping benchmarks TaskGrasp (parallel-jaw gripper) and DexGraspNet (Allegro and Shadow hands) demonstrate that GraspDreamer achieves superior data efficiency and generalization performance compared to prior methods. Real-world evaluations of the Allegro hand and parallel-jaw gripper further validate the effectiveness on real robots. Additionally, we showcase that GraspDreamer can be (1) naturally extended to downstream manipulation tasks and (2)  serve as an efficient data generation mechanism for visuomotor policy learning.





%% file: related.tex
\section{Related Work}\label{related}

\subsection{Functional Grasping}
 Prior work on functional grasping can be naturally organized from a data perspective. A first line of research is simulation-first synthesis, where large-scale datasets are generated from 3D assets and physics-based evaluation, and then used to train functional grasping policies \cite{xu2023unidexgrasp, he2025dexvlg, tang2023graspgpt, tang2025foundationgrasp, weng2024dexdiffuser, deshpande2025graspmolmo}. However, such synthetic supervision suffers from the sim2real gap and may not capture the functionality naturally expressed in human behavior, leading to physically stable grasps that are not aligned with downstream interaction objectives. To further incorporate human priors, a second line leverages human-first hand–object interaction (HOI) datasets \cite{chao2021dexycb, yang2022oakink, liu2024taco}, where functional intent is implicitly captured by human demonstrations and annotations such as affordance regions \cite{agarwaldexterous, wei2024grasp} and semantic touch/contact cues \cite{zhu2021toward}. In contrast to previous works, GraspDreamer generates human demonstrations with VGMs without labor-intensive data collection. While concurrent studies \cite{ye2025textsc, wei2025omnidexgrasp, patel2025robotic} also imitate functional grasping from generated images or videos, GraspDreamer further provides a unified framework that adapts across different robot hands.

    
\subsection{Imitating from Human Demonstrations}
Imitating grasping and manipulation skills from human demonstrations is widely viewed as a natural and scalable way to teach robots. On the grasping side, RTAGrasp \cite{dong2025rtagrasp} and HGDiffuser\cite{huang2025hgdiffuser} extract behavioral cues from in-the-wild images/videos for functional grasp synthesis. On manipulation, \cite{heppert2024ditto, liokami, tang2025mimicfunc} recover object-centric interaction trajectories from human videos and retarget them to robots, enabling spatial generalization \cite{heppert2024ditto, liokami} and functional generalization \cite{tang2025mimicfunc}. 
Beyond explicit trajectory representations, another line of work learns actionable visual affordance representations from human videos \cite{bahl2023affordances, mendonca2023structured} to guide downstream manipulation. More recently, EgoVLA \cite{yang2025egovla} combines robot and human videos for co-training vision-language-action (VLA) policies. Our work follows a similar spirit but differs in two key aspects.  First, GraspDreamer avoids large-scale data collection, enabling efficient and generalizable robot control. Second, while LVP \cite{chen2025large} shares a similar pipeline, GraspDreamer explicitly incorporates functionality and contact constraints during retargeting for improved performance.

\subsection{Generative Models for Robotics}
Generative models have recently emerged as a promising tool for robotics. Early work leverages LLMs for high-level planning over language-based plans \cite{ahn2022can} or structured graph representations \cite{ranasayplan}. Building on this idea of generative goal specification, prior works synthesize goal states with foundation generative models for object rearrangement \cite{kapelyukh2023dall} and dexterous grasping \cite{wei2025omnidexgrasp}. More recent works further adapt video foundation models into VLA policies \cite{kim2026cosmos, pai2025mimic}, demonstrating strong sample efficiency and generalization. GraspDreamer follows this trend by leveraging VGMs to generate human demonstrations in a zero-shot manner and transfer them across different robot hands for generalizable functional grasping.

%% file: approach.tex
\section{Approach}

\subsection{Problem Formulation}
Given the following inputs: an RGB-D scene observation $o \in \mathbb{R}^{H \times W \times 4}$, a language instruction $l$ specifying the task intent (e.g., ``\textit{grasp the handle of the knife}'' or ``\textit{handover the knife to me}''), and a URDF description $d_{\text{eef}}$ of the robot end-effector, the objective is to map the inputs to a grasp plan $\tau = \{ g_t \}_{t=0}^{T-1}$, which may include pre-grasp, grasp, and post-grasp phases, that satisfies the task intent. Directly predicting a grasp pose from a goal grasping image can be considered a special case corresponding to $T=1$. Formally, this mapping can be formulated as:
\begin{align*}
    \pi(o, l, d_{\text{eef}}) \rightarrow \tau
\end{align*}
where $g_t = (\mathbf{R}_t, \mathbf{t}_t,  \mathbf{q}_t) \in \text{SE(3)} \times \mathbb{R}^n$ represents the end-effector state at timestep $t$. Here, $\mathbf{R}_t \in \text{SO(3)}$ denotes the 3D orientation, $\mathbf{t}_t \in \mathbb{R}^3$ represents the 3D translation, and $\mathbf{q}_t \in \mathbb{R}^n$ corresponds to the joint configuration of the end-effector, where $n$ is the number of degrees of freedom. 

\begin{figure*}[t]
  \centering
  \vspace*{-0.1in}
  \begin{tikzpicture}[inner sep = 0pt, outer sep = 0pt]
    \node[anchor=south west] (fnC) at (0in,0in)
      {\includegraphics[height=4in,clip=true,trim=0.1in 0in 0in 0in]{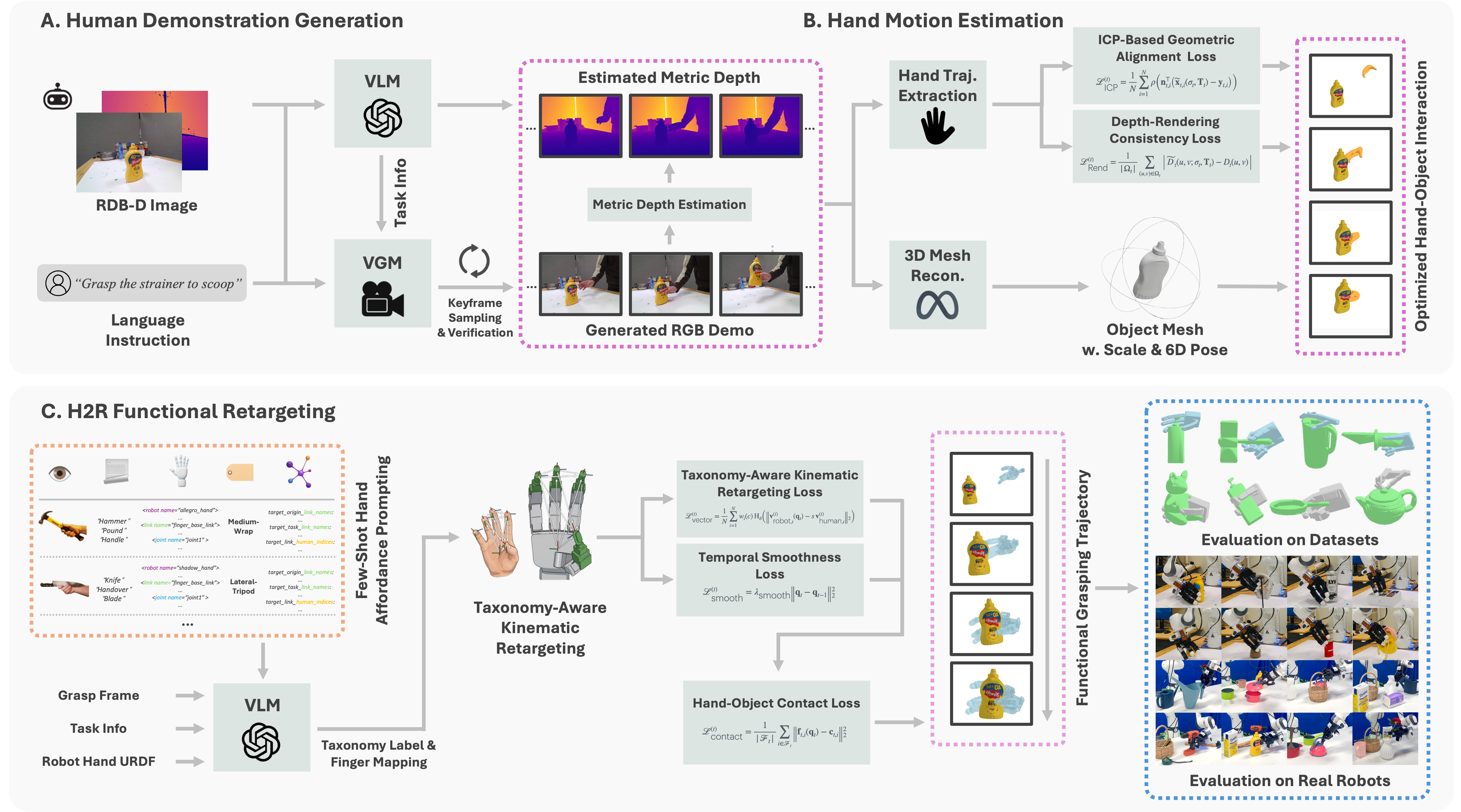}};
  \end{tikzpicture}
        \caption{An overview of GraspDreamer. The pipeline consists of three stages: (a) Human demonstration generation with VGMs, (b) Human hand motion extraction and optimization, and (c) Human-to-Robot functional retargeting and execution.}

  \label{fig:pipeline}
  \vspace*{-0.2in}
\end{figure*}

To address the formulated problem, we introduce GraspDreamer, a VGM-based framework for zero-shot functional grasping. An overview of the pipeline is shown in Figure~\ref{fig:pipeline}. Specifically, we first describe the human demonstration generation in Section~\ref{sec:demo_gen}. Section~\ref{sec: motion} then explains human hand motion estimation, and Section~\ref{sec: retarget} describes how the human motion is functionally retargeted to robot hands for functional grasping.


 \subsection{Human Demonstration Generation}\label{sec:demo_gen}

\textbf{Task-Relevant Information Extraction.} Due to the inherent ambiguity of natural language, the user-issued instruction may not explicitly specify task-relevant details, which can lead to suboptimal generation in the subsequent stage. To address this, we employ a VLM to reason about task-relevant information following a ``task-object-part" sequence. First, the VLM is prompted with $o$ and $l$ to output a task label $\mathcal{T}$. Second, given $\mathcal{T}$, the VLM identifies the most relevant object $\mathcal{O}$ in the scene that affords $\mathcal{T}$ and outputs a part decomposition $\mathcal{P} = \{ p_0, p_1, ..., p_m  \}$, where each $p_i$ denotes a part label. Finally, conditioned on $\mathcal{O}$ and $\mathcal{T}$, the VLM selects the part $p_\mathcal{T}$ from $\mathcal{P}$. Empirically, making $\mathcal{O}$, $\mathcal{P}$, and $\mathcal{T}$ explicit regularizes the subsequent generation.





\textbf{Visual Demonstration Generation.} With the extracted task-relevant information, GraspDreamer ``imagines" an RGB-D human demonstration $\mathcal{V}_H = \{I_t\}_{t=0}^{N-1}$ to guide downstream grasping, where each frame $I_t \in \mathbb{R}^{H \times W \times 4}$. In the first stage, given the visual observation $o$ and task-relevant information, we construct a prompt using a fixed-format template: ``\textit{A human hand grasps the $\mathcal{P}$ of $\mathcal{O}$ to $\mathcal{T}$},” along with the original $l$. Compared to recent works \cite{zhanggenerative, jang2025dreamgen} that fine-tune video generation or world models to roll out robot-centric videos, human demonstrations serve as a universal intermediate representation, which can be flexibly retargeted to diverse robot embodiments.  To mitigate hallucinations, we sample keyframes uniformly from the generated demonstration and send them to the VLM for verification. This closed-loop process continues until the VLM determines that the demonstration is consistent with the task intent.

In the second stage, we utilize the Video-Depth-Anything (VDA) model \cite{chen2025video} to predict the metric depth for each frame. While VDA provides temporally consistent depth across frames, the inherent scale ambiguity of monocular depth estimation results in insufficient precision for grasping. To tackle this, we leverage the target object depth information from $o$ for calibration. Specifically, we reconstruct the target object point cloud using both the true depth and the predicted depth, and estimate their rigid alignment with a weighted Umeyama algorithm \cite{umeyama2002least}. We then apply the resulting transform to all frames to calibrate the depth maps and obtain $\mathcal{V}_H$.

\subsection{Hand Motion Estimation} \label{sec: motion}

\textbf{Hand Trajectory Extraction.} With the generated human demonstration $\mathcal{V}_H$, we first use HaMeR \cite{pavlakos2024reconstructing} to estimate the hand state from each generated frame. Specifically, for the \(t\)-th frame, HaMeR outputs \((\mathbf{R}_t^h, \mathbf{t}_t^h, \theta_t^h)\), where \(\mathbf{R}_t^h \in \mathrm{SO}(3)\) and \(\mathbf{t}_t^h \in \mathbb{R}^3\) represent the 6 DoF wrist pose in the camera frame, and \(\theta_t^h \in \mathbb{R}^{45}\) denotes the MANO \cite{romero2017embodied} articulated hand pose parameters. The corresponding 3D hand joint locations, ${\mathbf{J}}_t^h \in \mathbb{R}^{21 \times 3}$, are then derived via the MANO layer. The object mesh can be optionally reconstructed with off-the-shelf models \cite{chen2025sam, xu2024instantmesh}. We apply RANSAC and ICP to estimate both the scale and transformation of the reconstructed object in each frame.

\textbf{Hand Trajectory Optimization.} A key challenge here is that VGMs lack metric-scale awareness; therefore, the hand–object size ratio in $\mathcal{V}_H$ may not align with real-world geometry. Consequently, directly fitting a fixed MANO template with an average human-hand scale often results in scale mismatch and depth-axis drift. To mitigate this, we optimize a per-frame scale factor $\sigma_t^{\text{hand}}$ for global correction and a 6D rigid transform $\mathbf{T}_t^{\text{hand}}$ for local refinement. Let $H_t$ denote the observed hand point cloud extracted from frame $I_t$, and
$H_t^{\text{mano}}$ the point cloud sampled from the MANO mesh. We consider an ICP-based geometric alignment term and a depth-rendering consistency term:
\subsubsection{ICP-Based Geometric Alignment.}
We minimize an ICP objective between the transformed MANO point cloud $\widetilde{H}_t^{\text{mano}}$ and the observed hand point cloud $H_t$:
\begin{equation*}
\mathcal{L}^{(t)}_{\text{ICP}}
=
\frac{1}{N_{\text{hand}}}
\sum_{i=1}^{N_{\text{hand}}}
\rho\!\left(
\mathbf{n}_{t,i}^{\top}
\bigl(
\widetilde{\mathbf{x}}_{t,i}(\sigma_t^{\text{hand}},\mathbf{T}_t^{\text{hand}})
-
\mathbf{y}_{t,i}
\bigr)
\right)
\label{eq:icp_loss}
\end{equation*}
where $\widetilde{\mathbf{x}}_{t,i}\in \widetilde{H}_t^{\text{mano}}$ is the
$i$-th transformed MANO point, $\mathbf{y}_{t,i}\in H_t$ is its current
correspondence (updated iteratively following standard ICP), and
$\mathbf{n}_{t,i}$ is the surface normal at $\mathbf{y}_{t,i}$. The penalty $\rho(\cdot)$ (e.g., Huber) mitigates the influence of noisy depth and outliers.

\subsubsection{Depth-Rendering Consistency.}
To anchor the alignment in image space, we rasterize the transformed
MANO mesh to obtain a rendered depth map
$\widetilde{D}_t(\sigma_t^{\text{hand}},\mathbf{T}_t^{\text{hand}})$ and compare it to the observed
hand depth $D_t$:
\begin{equation*}
\mathcal{L}^{(t)}_{\text{rend}}
=
\frac{1}{|\Omega_t|}
\sum_{(u,v)\in\Omega_t}
\big|
\widetilde{D}_t(u,v;\sigma_t^{\text{hand}},\mathbf{T}_t^{\text{hand}})
-
D_t(u,v)
\big|
\label{eq:render_loss}
\end{equation*}
where $\Omega_t$ denotes pixels belonging to the hand region. Since $\mathbf{T}_t^{\text{hand}}$ is only intended to be a local refinement, we also regularize it to remain close to the identity transform.



\subsection{H2R Functional Retargeting} \label{sec: retarget}

\textbf{Hand Affordance Reasoning.} To enable functional H2R retargeting, GraspDreamer first reasons the hand affordances for grasping. These affordances are taxonomy-dependent, specifying both digit roles (which digits establish task-relevant contacts) and coupling patterns (how inter-finger coordination realizes the intended grasp function). Accordingly, functional retargeting requires (i) a finger mapping to translate human digit roles to the target robot hand, and (ii) an inferred grasp taxonomy to guide how these roles should be emphasized during retargeting. Unlike prior work that assumes a fixed manual mapping \cite{handa2020dexpilot}, we exploit VLM in-context learning to infer task-specific taxonomy and finger mappings. We build a few-shot prompt with grasp images, object/task/part labels (as in Section ~\ref{sec:demo_gen}), taxonomy labels, robot hand URDFs, and human-defined optimal mappings. Following \cite{li2025language}, we use 12 basic grasp types from \cite{feix2015grasp}, and at inference time, the VLM predicts both the taxonomy and the mapping conditioned on the generated grasp frame, task cues, and the URDF.



\textbf{Taxonomy-Aware Kinematic Retargeting.} In the second stage, we incorporate predicted hand affordances into the retargeting process.   For each frame $t$, let $\mathbf{q}_t$ denote the robot joint configuration, 
$\mathbf{v}_{\text{robot},t}^{(i)}(\mathbf{q}_t)$ the $i$-th robot vector computed via forward kinematics, 
and $\mathbf{v}_{\text{ref},t}^{(i)}$ the corresponding reference vector derived from the human hand joints $\textbf{J}_t$.  
Given a taxonomy class $c$, the taxonomy-aware vector matching loss for frame $t$ is defined as
\begin{equation*}
\mathcal{L}^{(t)}_{\text{vector}}
=
\frac{1}{N_{\text{vec}}}
\sum_{i=1}^{N_{\text{vec}}}
w_i(c)\,
\mathrm{H}_{\delta}
\Bigl(
\bigl\|
\mathbf{v}_{\text{robot},t}^{(i)}(\mathbf{q}_t)
-
s\,\mathbf{v}_{\text{ref},t}^{(i)}
\bigr\|_2
\Bigr)
\label{eq:tax_vector_loss}
\end{equation*}
where $N_{\text{vec}}$ is the number of vector pairs, $s$ is a global scale factor, and $\mathrm{H}_{\delta}(\cdot)$ denotes  the Huber loss with threshold $\delta$. $w_i(c)$ is a taxonomy-dependent weight. For example, in a medium-wrap grasp, such as using a hammer to pound, vectors related to global enclosure around the handle receive higher weights. In contrast, in a lateral-tripod grasp such as handing over a knife, the optimizer instead upweights thumb–index–middle relations that define the lateral pinch, while de-emphasizing the remaining fingers. To ensure temporal smoothness, we additionally penalize deviations from the previous
configuration:
\begin{equation*}
\mathcal{L}^{(t)}_{\text{smooth}}
=
\lambda_{\text{smooth}}
\bigl\|
\mathbf{q}_t - \mathbf{q}_{t-1}
\bigr\|_2^2
\end{equation*}
The overall taxonomy-aware retargeting objective is thus:
\begin{equation*}
\mathcal{L}_{\text{retarget}}^{(t)}
=
\mathcal{L}^{(t)}_{\text{vector}}
+
\mathcal{L}^{(t)}_{\text{smooth}}
\label{eq:full_tax_objective}
\end{equation*}
subject to joint limits
$\mathbf{q}_{\min} \le \mathbf{q}_t \le \mathbf{q}_{\max}$. We solve this constrained optimization for each frame $t$ using Sequential Least Squares Programming (SLSQP).



\textbf{Hand-Object Contact Refinement.} For each contact frame, we further refine the hand configuration by alternating between optimizing the articulated joint configuration and a rigid correction of the wrist pose. Let $\mathcal{F}_{t,\text{aff}}$ denote the set of visible fingertips whose projections fall within the task-relevant region. From $\mathcal{V}_H$, we extract the corresponding 3D human fingertip points $\mathcal{K}_t=\{\mathbf{c}_{t,i}\}\subset\mathbb{R}^3$ on the same region. For each fingertip $i\in\mathcal{F}_{t,\text{aff}}$, let $\mathbf{f}_{t,i}(\mathbf{q}_t)\in\mathbb{R}^3$ denote its 3D position under forward kinematics from the robot joint configuration $\mathbf{q}_t$. We then update $\mathbf{q}_t$ by minimizing the hand-object contact loss:
\[
\mathcal{L}^{(t)}_{\text{contact}}
=
\frac{1}{|\mathcal{F}_{t,\text{aff}}|}
\sum_{i \in \mathcal{F}_{t,\text{aff}}}
\bigl\|
\mathbf{f}_{t,i}(\mathbf{q}_t)
-
\mathbf{c}_{t,i}
\bigr\|_2^2
\]
The rigid pose refinement follows a similar formulation and is omitted here for brevity. To ensure temporal smoothness and prevent over-correction, we also regularize both the deviation from the initial retargeted joints. This objective yields physically plausible fingertip contacts while maintaining the intended grasp functionality. For parallel-jaw grasping, we empirically observed that applying contact refinement based on \cite{sundermeyer2021contact} as in \cite{dong2025rtagrasp, tang2023graspgpt, tang2025foundationgrasp} results in improved performance.

%% file: exp_setup.tex
\section{Experimental Setup}\label{exp_setup}

\subsection{Simulation Experiments}
\textbf{Baselines.} We compare GraspDreamer against parallel-jaw and dexterous grasping baselines. 
For parallel-jaw functional grasping, we include a training-free method LAN-Grasp~\cite{mirjalili2023lan} and three training-based methods: GraspGPT~\cite{tang2023graspgpt}, FoundationGrasp~\cite{tang2025foundationgrasp}, and GraspMolmo~\cite{deshpande2025graspmolmo}. For dexterous grasping, we compare with two hand-specific baselines, DexGYS~\cite{wei2024grasp} and DexDiffuser~\cite{weng2024dexdiffuser}. Since DexDiffuser does not natively support task specification, we augment it with part localization and segmentation for fair comparison.


\textbf{Benchmark Datasets.} For parallel-jaw grasping, we evaluate all methods on TaskGrasp~\cite{murali2021same} and report results on \texttt{t\_split\_0} and \texttt{o\_split\_0}, which test task and object generalization, respectively. For dexterous grasping, we evaluate on a subset of DexGraspNet~\cite{wang2023dexgraspnet} comprising 13 categories and 27 object instances with Allegro and Shadow hands, covering kitchenware and mechanical tools that require functional use beyond simple pick-and-place. Since DexGraspNet lacks task labels for ground-truth grasps, we manually annotate them following a procedure similar to that in \cite{murali2021same}. As both TaskGrasp and DexGraspNet are static-grasp datasets, we evaluate using only the generated grasp frame.

\textbf{Evaluation Metrics.} Following prior work~\cite{deshpande2025graspmolmo, tang2023graspgpt}, we use the top-1 success rate, termed \emph{Success}, as our primary evaluation metric. Additionally, we evaluate whether the robot selects the correct functional part for grasping, termed \emph{Part Identification Accuracy (PIA)}. Human experts further assess whether the predicted grasp aligns with the intended task using a Likert-scale rating, following \cite{tang2025foundationgrasp}. We refer to this metric as \emph{Intent}. The Likert scale is defined as: Strongly Agree ($5$), Agree ($4$), Neither Agree nor Disagree ($3$), Disagree ($2$), and Strongly Disagree ($1$). 


\subsection{Real-Robot Experiments}
Real-robot experiments are conducted on Franka Panda arms equipped with two end-effectors: a 16-DoF Allegro Hand and a Robotiq 2F-85 parallel-jaw gripper. An eye-to-hand calibrated RGB-D camera provides single-view observations. We evaluate performance by recording the success rate over 10 repeated trials per object. Test objects are drawn from both laboratory collections and the YCB dataset.

%% file: exp.tex
\section{Experiments}\label{exp}

\begin{table}[t]
\centering
\renewcommand\arraystretch{2}
\setlength\tabcolsep{3pt}
\begin{tabular}{ccccccc}
\toprule
\multirow{2}{*}{\textbf{Method}} & \multicolumn{3}{c}{\textbf{Object Generalization}} & \multicolumn{3}{c}{\textbf{Task Generalization}} \\ \cline{2-7} 
                                 & \emph{Success}          & \emph{PIA}              & \emph{Intent}      & \emph{Success}        & \emph{PIA}            & \emph{Intent}        \\ \hline
LAN-Grasp                        & 68.2\%            & 92.8\% & 1.9          & 65.3\%          & 88.5\%          & 2.0          \\
GraspGPT                         & 71.4\%           & 83.4\%          & 2.8          & 73.4\%          & 81.4\%          & 2.5          \\
FoundationGrasp                  & 73.5\%           & 86.8\%          & 2.9          & 74.2\%          & 82.0\%          & 2.8          \\
GraspMolmo                       & 77.4\%           & 89.1\%          & 3.3          & 76.7\%          & 90.1\% & 3.5          \\ \hline
GraspDreamer (Ours)              & 78.6\%  & 86.1\%          & 3.6 & 79.5\% & 85.2\%          & 3.8 \\ \bottomrule
\end{tabular}
\caption{Quantitative evaluation on the TaskGrasp dataset.}
\label{tab:taskgrasp}

\vspace{4mm}

\begin{tabular}{ccccccc}
\toprule
\multirow{2}{*}{\textbf{Method}} & \multicolumn{3}{c}{\textbf{Kitchenware}} & \multicolumn{3}{c}{\textbf{Mechanic Tool}} \\ \cline{2-7} 
                        & \emph{Success}     & \emph{PIA}      & \emph{Intent}   & \emph{Success}      & \emph{PIA}       & Intent   \\ \hline
DexGYS (Shadow)       & 56.2\%   & 55.4\%   &  2.3     &   56.3\%  &  60.3\%   &  2.6     \\ 
GraspDreamer (Shadow)   &  80.2\%  &  85.7\%  &   3.4    &  83.1\%   &  87.2\%    &    3.6   \\ 
DexDiffuser (Allegro)   &  49.3\%  &  82.4\% &   2.2    &  60.4\%    & 82.3\%   &   2.7    \\ 
GraspDreamer (Allegro)  & 78.5\%   & 84.3\%   &  3.3      &   82.6\%  &  85.4\%  & 3.5     \\ \bottomrule
\end{tabular}
\caption{Quantitative evaluation on the DexGraspNet dataset.}
\label{tab:dexgraspnet}
\vspace{-6mm}
\end{table}

\subsection{Results of Simulation Experiments}

\begin{figure}[t]
  \centering
  \begin{tikzpicture}[inner sep = 0pt, outer sep = 0pt]
    \node[anchor=south west] (fnC) at (0in,0in)
      {\includegraphics[height=1.5in,clip=true,trim=0in 0in 0in 0in]{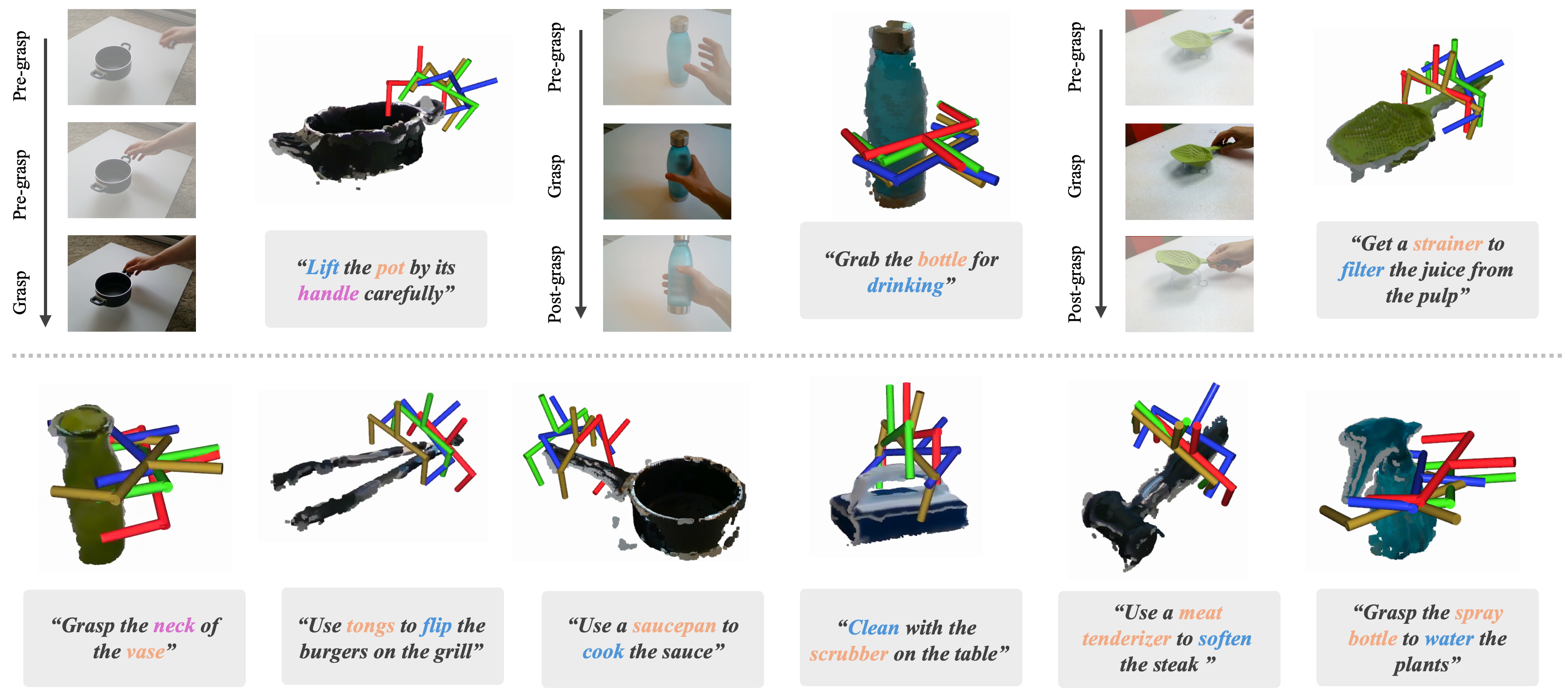}};
  \end{tikzpicture}
        \caption{Qualitative results on the TaskGrasp dataset. The predicted grasp is shown in blue, and the top-3 matched grasps from the dataset are visualized in green, brown, and red, respectively.}
      \label{fig:taskgrasp}

  \begin{tikzpicture}[inner sep = 0pt, outer sep = 0pt]
    \node[anchor=south west] (fnC) at (0in,0in)
      {\includegraphics[height=2in,clip=true,trim=0.2in 0in 0in 0in]{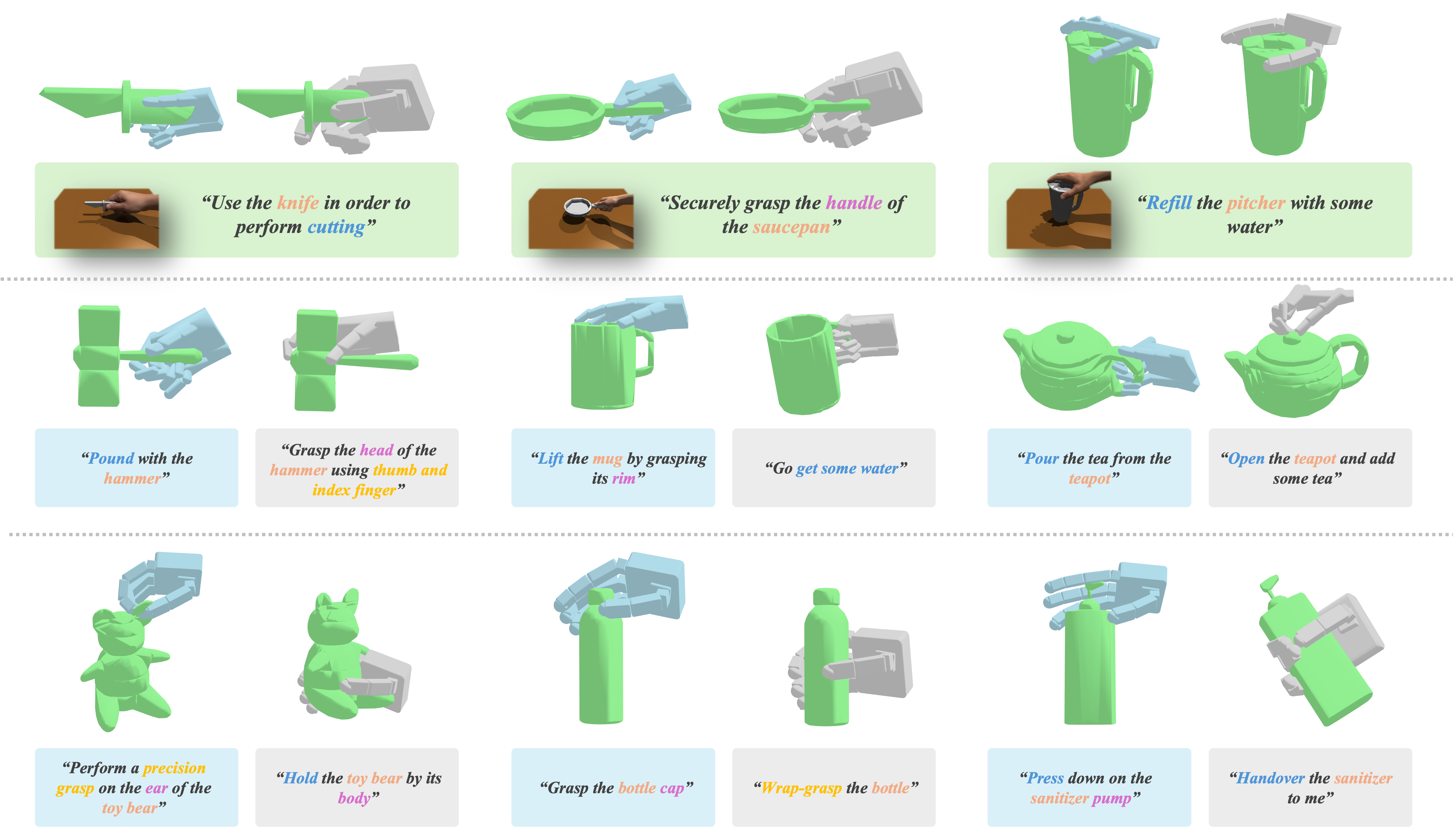}};
  \end{tikzpicture}
        \caption{Qualitative results on the DexGraspNet dataset. The first row shows the same task with different hands, while the remaining rows show the same object across different tasks using the Shadow and Allegro hands.}

  \label{fig:dexgraspnet}

  \vspace*{-6mm}
\end{figure}

\textbf{TaskGrasp Evaluation.} The quantitative results are shown in Table~\ref{tab:taskgrasp}. 
Compared to the training-free baseline LAN-Grasp, which focuses on functional region localization, GraspDreamer achieves higher \emph{Success} and \emph{Intent}, benefiting from generalizable interaction priors encoded in VGMs that inform both \emph{where} and \emph{how} to grasp. It also outperforms training-based baselines without requiring additional robot data collection. Notably, although GraspMolmo is adapted from Molmo, its behavior remains largely influenced by the functional grasping dataset used for fine-tuning. Additionally, several baselines obtain high \emph{PIA} yet fail to produce intent-aligned grasps, underscoring that functional grasping demands more than part localization. Qualitative results are shown in Figure~\ref{fig:taskgrasp}.


\textbf{DexGraspNet Evaluation.} For dexterous functional grasping, quantitative results are reported in Table~\ref{tab:dexgraspnet}. Although DexGYS is trained on a large language-conditioned dexterous grasp dataset, it primarily optimizes finger contacts to match language-specified constraints, which does not necessarily capture the functional intent. DexDiffuser can generate physically stable dexterous grasps, but it lacks an explicit mechanism to ground task functionality during grasp synthesis. As a result, it may achieve high \emph{PIA} while still producing grasps that are not functionally valid. Figure~\ref{fig:dexgraspnet} presents qualitative results obtained with both the Shadow and Allegro hands.

\begin{table}[h]
\centering
\renewcommand\arraystretch{1.5}
\setlength\tabcolsep{2pt}
\vspace{-1mm}
\begin{tabular}{cccc}
\toprule
\textbf{Object}       & \textbf{Success} (Gripper) & \textbf{Object}       & \textbf{Success} (Dex) \\ \hline
Watering Pot & 8/10              & Chip Can     & 7/10               \\
Box          & 7/10              & Toy          & 9/10               \\
Brush        & 6/10              & Box          & 8/10               \\
Mug          & 6/10              & Water Bottle & 7/10               \\
Power Drill  & 9/10              & Wine Glass   & 6/10               \\
Basket       & 7/10              & Milk Bottle  & 8/10               \\
Pot          & 9/10              & Bag          & 9/10               \\
Bottle       & 7/10              & Headphones   & 5/10               \\
Teapot       & 10/10             & Tennis Ball  & 8/10               \\
Spoon        & 7/10              & Red Block    & 6/10               \\ \hline
\textbf{Total}        & 76/100            & \textbf{Total}        & 73/100             \\ \bottomrule
\end{tabular}
\caption{Quantitative results of real-robot experiments: parallel-jaw grasping (left) and dexterous grasping (right).
}
\label{tab:grasp}
\vspace{-2mm}
\end{table}

\subsection{Results of Real-Robot Experiments}

\textbf{Functional Grasping.} Table~\ref{tab:grasp} reports real-robot results of GraspDreamer on both Robotiq 2F-85 (left) and Allegro hand (right). The test set comprises commonly used household objects with diverse functions, sizes, and geometries. GraspDreamer achieves success rates above 70\% in both settings. Notably, for interactions that require precise contact and force control (e.g., grasping the headphone headband), performance is lower, reflecting the increased sensitivity to small pose errors and contact dynamics.







\textbf{Extension to Manipulation.} We evaluate GraspDreamer on three short-horizon manipulation tasks by additionally extracting post-grasp trajectories from the generated demonstrations. Each trial is decomposed into three stages: (i) visual demonstration generation, (ii) grasping, and (iii) manipulation. As reported in Table~\ref{tab:mani}, GraspDreamer achieves relatively high success on \texttt{Pull Tissue} and \texttt{Take Flower from Vase}, while \texttt{Open Pot} remains more challenging due to tighter contact constraints and higher sensitivity to pose and force errors. The gap between the grasping and manipulation stages also highlights that VGMs become more prone to hallucination over longer horizons, often generating visually or physically infeasible post-grasp motions.

\textbf{Extension to Policy Learning.} We further demonstrate that GraspDreamer rollouts can be used to train visuomotor policies without requiring human teleoperation. Specifically, we collect 50 demonstrations generated by GraspDreamer for two tasks, \texttt{Pull Tissue} (from the box) and \texttt{Pick up Bottle}, and use them to train Diffusion Policies \cite{chi2025diffusion}. Despite being generated without manual intervention, these demonstrations provide sufficient interaction cues for policy learning. The resulting policies achieve success rates of 73.3\% and 86.7\%, respectively, indicating that the generated trajectories can serve as effective supervision for downstream manipulation learning. These results highlight the potential of generative human demonstrations as an alternative data source for training visuomotor policies.




\begin{table}[t]
\centering
\renewcommand\arraystretch{1.5}
\setlength\tabcolsep{1.5pt}
\begin{tabular}{ccccc}
\toprule
\multirow{2}{*}{\textbf{Task Description}} & \multicolumn{3}{c}{\textbf{Stage}}  & \textbf{Success}              \\ \cline{2-4}
                      & Generation & Grasp & Manipulation & \multicolumn{1}{l}{} \\ \hline
\texttt{Take Flower from Vase} &  9/10   &  7/10     &      7/10        &           70\%           \\
\texttt{Open Pot}          &  7/10  & 6/10      &    5/10          &              50\%        \\
\texttt{Pull Tissue}           &  8/10   &   8/10   &   8/10         &              80\%        \\ \bottomrule
\end{tabular}
\caption{Quantitative evaluation on three short-horizon manipulation tasks.}
\label{tab:mani}
  \vspace*{-6mm}
\end{table}

\subsection{Ablation Studies}
\begin{figure}[h]
  \centering
  \vspace*{-0.1in}
  \begin{tikzpicture}[inner sep = 0pt, outer sep = 0pt]
    \node[anchor=south west] (fnC) at (0in,0in)
      {\includegraphics[height=0.7in,clip=true,trim=0in 0in 0in 0in]{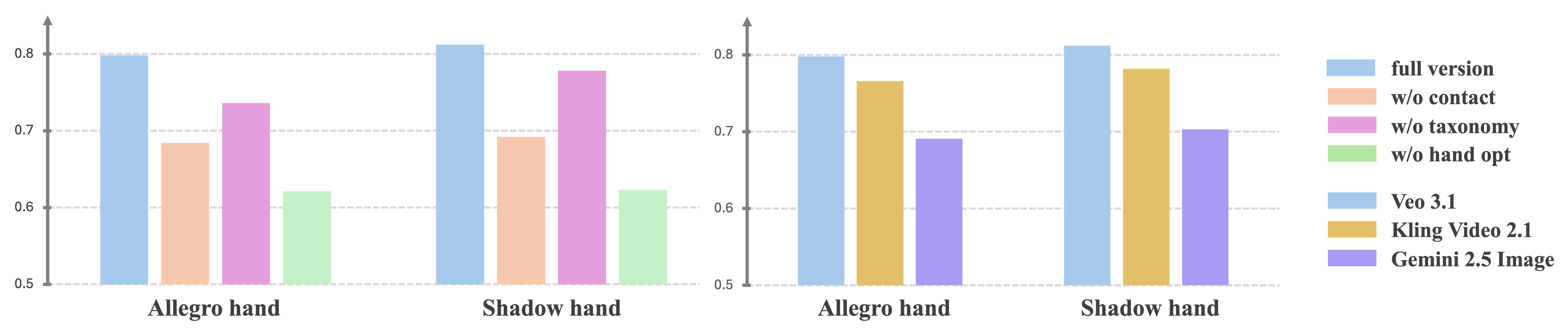}};
  \end{tikzpicture}
        \caption{Ablation studies on system components (left) and generation models (right).}
  \label{fig:ablation}
\end{figure}

\textbf{Ablation on System Components.} To assess the role of each component in GraspDreamer, we perform an ablation study on DexGraspNet with three variants: (i) w/o contact, removing hand–object contact refinement; (ii) w/o taxonomy, removing taxonomy-aware retargeting constraints; and (iii) w/o hand opt, removing hand trajectory optimization. As shown in Figure~\ref{fig:ablation} (left), the trends are consistent across both dexterous hands. Directly using wrist poses from generated demonstrations without optimization yields the worst results. Moreover, because of the morphological gap between human and robot hands, contact refinement significantly improves contact accuracy and grasp feasibility.

\textbf{Ablation on Generative Models.} We further compare three VGMs: two video models (Veo 3.1 and Kling Video 2.1) and an image model (Gemini 2.5 Image). Empirically, video generation produces more temporally coherent interactions and physically plausible motion, leading to better downstream grasp performance than independently generated images. Among the video models, Veo 3.1 slightly outperforms Kling Video 2.1, as shown in Figure~\ref{fig:ablation} (right).

%% file: discussion.tex
\section{Conclusion}\label{conclusion}
In this paper, we present GraspDreamer, which leverages human demonstrations generated by VGMs to enable zero-shot functional grasping without labor-intensive robot data collection. Compared to prior methods, GraspDreamer achieves improved data efficiency and generalization on public benchmarks and can be flexibly adapted across different robot hands. Real-world evaluations further validate its effectiveness on real robots. Additionally, we demonstrate that GraspDreamer can (1) be naturally extended to downstream manipulation tasks and (2) provide training data for visuomotor policy learning. In future work, we aim to further develop GraspDreamer into a scalable and efficient framework for large-scale data generation in functional grasping and manipulation.